\documentclass[10pt,twocolumn,letterpaper]{article}

\usepackage[pagenumbers]{cvpr} %

\usepackage{makecell}
\usepackage{multirow}

\definecolor{nbarrier}{RGB}{255, 120, 50}
\definecolor{nbicycle}{RGB}{255, 192, 203}
\definecolor{nbus}{RGB}{255, 255, 0}
\definecolor{ncar}{RGB}{0, 150, 245}
\definecolor{nconstruct}{RGB}{0, 255, 255}
\definecolor{nmotor}{RGB}{200, 180, 0}
\definecolor{npedestrian}{RGB}{255, 0, 0}
\definecolor{ntraffic}{RGB}{255, 240, 150}
\definecolor{ntrailer}{RGB}{135, 60, 0}
\definecolor{ntruck}{RGB}{160, 32, 240}
\definecolor{ndriveable}{RGB}{255, 0, 255}
\definecolor{nother}{RGB}{139, 137, 137}
\definecolor{nsidewalk}{RGB}{75, 0, 75}
\definecolor{nterrain}{RGB}{150, 240, 80}
\definecolor{nmanmade}{RGB}{213, 213, 213}
\definecolor{nvegetation}{RGB}{0, 175, 0}

\definecolor{nvcolor}{RGB}{119,185,0}
\definecolor{roadcolor}{RGB}{234,51,246}
\definecolor{sidewalkcolor}{RGB}{68,8,72}
\definecolor{parkingcolor}{RGB}{241,156,249}
\definecolor{othergroundcolor}{RGB}{160,32,76}
\definecolor{buildingcolor}{RGB}{246,202,69}
\definecolor{carcolor}{RGB}{111,149,238}
\definecolor{truckcolor}{RGB}{74,32,172}
\definecolor{bicyclecolor}{RGB}{136,227,242}
\definecolor{motorcyclecolor}{RGB}{37,59,146}
\definecolor{othervehiclecolor}{RGB}{96,81,242}
\definecolor{vegetationcolor}{RGB}{79, 173, 50}
\definecolor{trunkcolor}{RGB}{126, 65, 22}
\definecolor{terraincolor}{RGB}{171, 238, 105}
\definecolor{personcolor}{RGB}{234, 60, 49}
\definecolor{bicyclistcolor}{RGB}{234, 66, 195}
\definecolor{motorcyclistcolor}{RGB}{138, 42, 90}
\definecolor{fencecolor}{RGB}{238, 128, 69}
\definecolor{polecolor}{RGB}{252, 241, 161}
\definecolor{trafficsigncolor}{RGB}{233, 51, 35}
\definecolor{other-struct.color}{RGB}{255, 150, 0}
\definecolor{other-objectcolor}{RGB}{50, 255, 255}
\definecolor{lane-markingcolor}{RGB}{150, 255, 170}
\definecolor{color1}{RGB}{176, 36, 24}
\definecolor{color2}{RGB}{0, 176, 80}
\definecolor{color3}{RGB}{0, 0, 200}

\definecolor{cvprblue}{rgb}{0.21,0.49,0.74}
\usepackage[pagebackref,breaklinks,colorlinks,allcolors=cvprblue]{hyperref}

\def\paperID{*****} %
\def\confName{CVPR}
\def\confYear{2025}

\title{GaussianFormer-2: Probabilistic Gaussian Superposition \\ for Efficient 3D Occupancy Prediction}

\author{Yuanhui Huang$^1$ \quad Amonnut Thammatadatrakoon$^1$ \quad Wenzhao Zheng$^{1,}$\footnotemark[1] \\ Yunpeng Zhang$^2$ \quad Dalong Du$^{1,2}$ \quad Jiwen Lu$^1$ \\
$^1$Tsinghua University, China \quad $^2$PhiGent Robotics \\
\texttt{\{huangyh22,yam23\}@mails.tsinghua.edu.cn; wenzhao.zheng@outlook.com} 
}

\begin{document}

\twocolumn[{%
\renewcommand\twocolumn[1][]{#1}%
\vspace{-20mm}
\maketitle
\vspace{-10mm}
\begin{center}
    \centering
    \includegraphics[width=\linewidth]{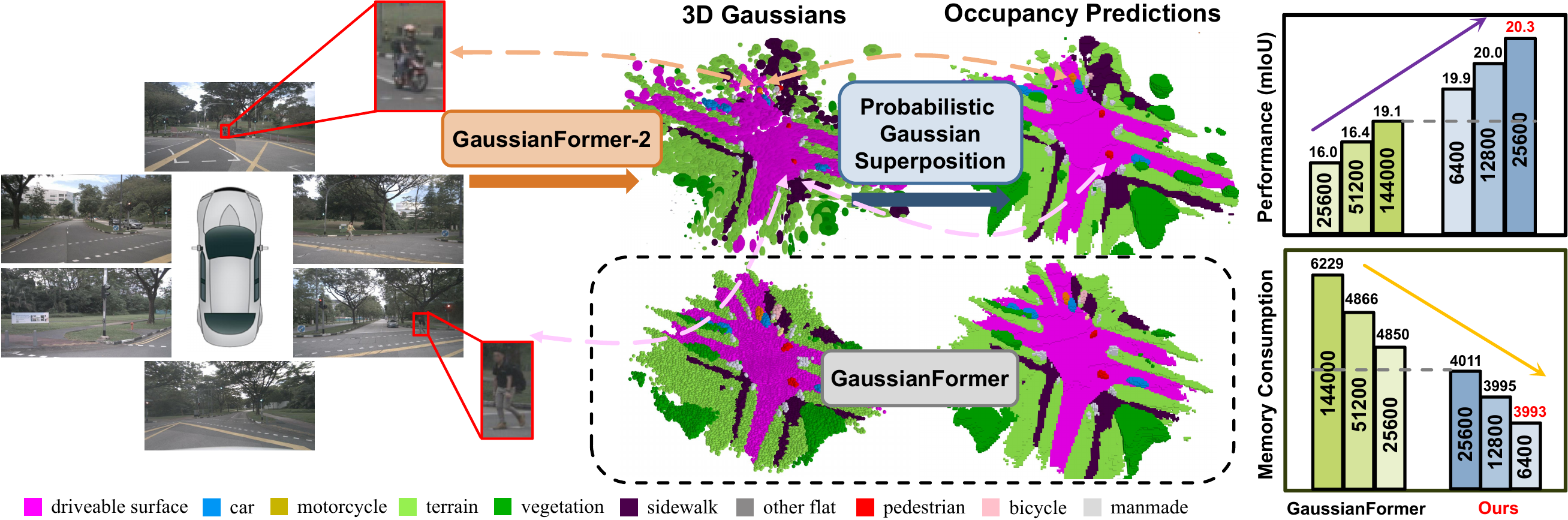}
    \vspace{-7mm}
    \captionof{figure}{
    We approach efficient object-centric scene representation from a probabilistic perspective and propose the probabilistic Gaussian superposition model, which achieves SOTA performance with as few as 8.9\% of Gaussians in GaussianFormer~\cite{huang2024gaussian}.
    }
    \vspace{-1mm}
\label{teaser}
\end{center}%
}]

\renewcommand{\thefootnote}{\fnsymbol{footnote}}
\footnotetext[1]{Project leader.}
\renewcommand{\thefootnote}{\arabic{footnote}}

\begin{abstract}
3D semantic occupancy prediction is an important task for robust vision-centric autonomous driving, which predicts fine-grained geometry and semantics of the surrounding scene. 
Most existing methods leverage dense grid-based scene representations, overlooking the spatial sparsity of the driving scenes.
Although 3D semantic Gaussian serves as an object-centric sparse alternative, most of the Gaussians still describe the empty region with low efficiency.
To address this, we propose a probabilistic Gaussian superposition model which interprets each Gaussian as a probability distribution of its neighborhood being occupied and conforms to probabilistic multiplication to derive the overall geometry.
Furthermore, we adopt the exact Gaussian mixture model for semantics calculation to avoid unnecessary overlapping of Gaussians.
To effectively initialize Gaussians in non-empty region, we design a distribution-based initialization module which learns the pixel-aligned occupancy distribution instead of the depth of surfaces.
We conduct extensive experiments on nuScenes and KITTI-360 datasets and our GaussianFormer-2 achieves state-of-the-art performance with high efficiency.
Code: \url{https://github.com/huang-yh/GaussianFormer}.
\end{abstract}

\vspace{-3mm}
\section{Introduction}
\label{sec:intro}
In autonomous driving, vision-centric systems have been more cost-effective compared with the LiDAR-based counterparts.
However, their inability to capture obstacles with arbitrary shapes poses challenges for driving safety and reliability~\cite{li2022bevformer,hu2022uniad,jiang2023vad,li2022bevdepth}. 
The advent of 3D semantic occupancy prediction methods~\cite{cao2022monoscene,miao2023occdepth,zhang2023occformer,wei2023surroundocc,jiang2023symphonize,huang2023tri,li2023voxformer,li2023fb} alleviates this limitation by predicting the fine-grained geometry and semantics of the surrounding 3D environment. 
This advancement supports a range of emerging applications, including end-to-end autonomous driving~\cite{occnet,hu2022uniad}, 4D occupancy forecasting~\cite{occworld,wang2024occsora,yan2024renderworld}, and self-supervised 3D scene understanding~\cite{selfocc,wimbauer2023behind,cao2023scenerf}.

Despite the promising applications, 3D semantic occupancy prediction is essentially a dense three-dimensional segmentation task~\cite{cao2022monoscene,tian2023occ3d}, which necessitates a both efficient and effective representation of the 3D scene.
Voxel-based methods~\cite{li2023voxformer,wei2023surroundocc} use dense 3D voxels as representation to describe the scene with the finest detail.
However, they neglect the spatial redundancy of the 3D occupancy and suffer from high computational complexity.
As a workaround, planer representations, such as BEV~\cite{li2022bevformer,yu2023flashocc} and TPV~\cite{huang2023tri}, compress the 3D grid along one of the axes to derive 2D feature maps for reduction of the token number.
Nonetheless, they still take into account the empty region when modeling the environment, which compromises their model capacity and efficiency.
As a pioneer in object-centric sparse scene representations, 3D semantic Gaussians~\cite{huang2024gaussian} describe the 3D space in a sparse way with learnable mean, covariance, opacity and semantics for each Gaussian.
However, several limitations persist in the current 3D semantic Gaussian representation:
1) Each Gaussian can still describe the empty region, which renders most of the Gaussians useless in an object-centric formulation given the sptial sparsity of 3D occupancy.
2) The aggregation process ignores the overlapping issue and directly sums up the contribution of each Gaussian to produce occupancy prediction, which results in unbounded semantic logits and further increases the overlapping among Gaussians.
Thus, the proportion of effective Gaussians describing occupied regions independently could be extremely low, which undermines the efficiency of the 3D semantic Gaussian representation.

In this paper, we introduce a probabilistic Gaussian superposition model to resolve the above limitations of 3D semantic Gaussians and improve utilization and efficiency. 
To elaborate, we propose the probabilistic Gaussian representation, which assigns 3D Gaussians to exclusively model the non-empty area by interpreting each Gaussian as a probability distribution of its neighborhood being occupied.
We employ the multiplication theorem of probability to aggregate the independent probability distributions and derive the geometry predictions.
Furthermore, we integrate the Gaussian mixture model into out probabilistic Gaussian representation to generate normalized semantic predictions, which avoid unbounded logits and prevents Gaussians from unnecessary overlapping.
Since our representation only models the occupied region, we also design a distribution-based initialization module to effectively initialize Gaussians around the non-emtpy area, which learns the pixel-aligned occupancy distribution instead of depth values of surfaces~\cite{li2023voxformer,li2022bevdepth,huang2021bevdet}.
We conduct extensive experiments on the nuScenes~\cite{caesar2020nuscenes} and KITTI-360~\cite{Liao2022kitti360} datasets for surround-view and monocular 3D semantic occupancy prediction, respectively. 
Our GaussianFormer-2 outperforms state-of-the-art methods
with high efficiency. 
In addition, qualitative visualizations show that GaussianFormer-2 is able to generate a both holistic and realistic perception of the scene.

\section{Related Work}
\label{sec: related work}

\begin{figure}[t]
\centering
\includegraphics[width=0.95\linewidth]{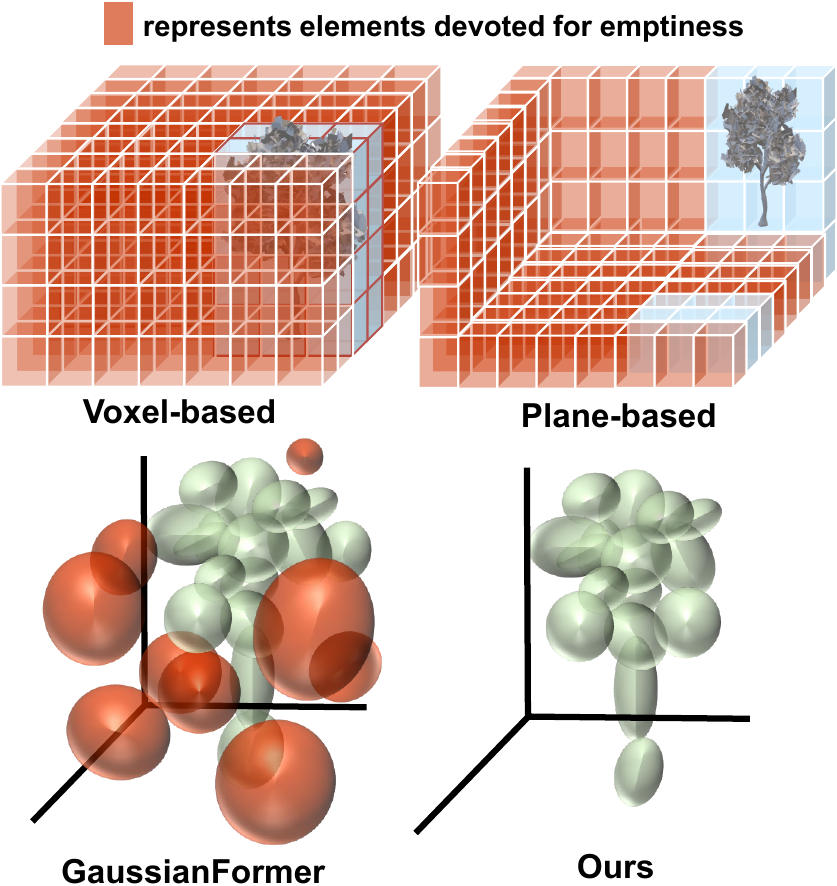}
\vspace{-2mm}
\caption{\textbf{Representation comparisons.}
Voxel and plane based representations inevitably incorporate emptiness when modeling the 3D scene.
While GaussianFormer~\cite{huang2024gaussian} proposes 3D semantic Gaussian as a sparse representation, it still suffer from spatial redundancy.
Our method achieves true object-centricity through probabilistic modeling.
}
\label{fig:motivation}
\vspace{-7mm}
\end{figure}

\begin{figure*}[t]
\centering
\includegraphics[width=0.95\linewidth]{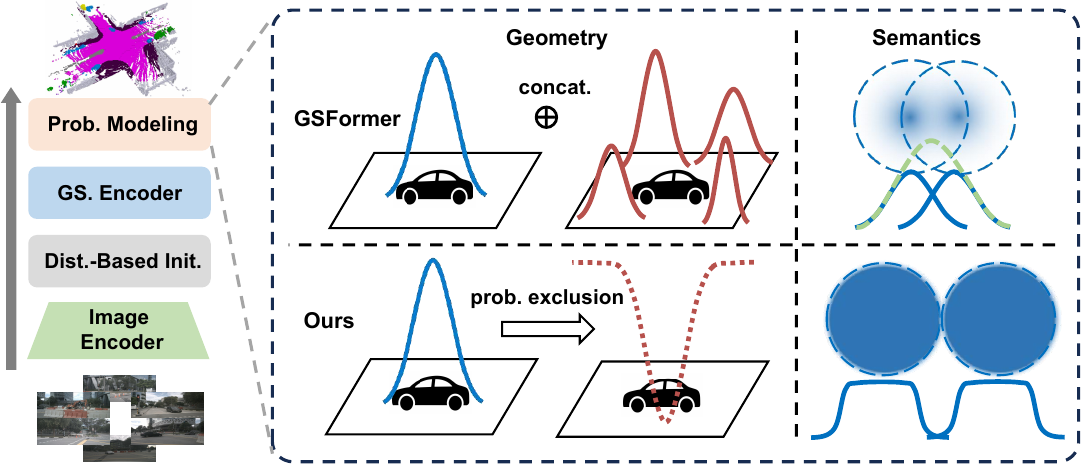}
\vspace{-2mm}
\caption{\textbf{Overall pipeline of our method.}
To achieve probabilistic modeling, we decompose occupancy prediction into geometry and semantics prediction, and approach them separately using probabilistic multiplication and Gaussian mixture model to improve efficiency.
}
\label{fig:pipeline}
\vspace{-6mm}
\end{figure*}

\textbf{3D Semantic Occupancy Prediction.}
3D semantic occupancy prediction~\cite{cao2022monoscene,huang2023tri,tian2023occ3d} has emerged as a promising environment modeling in autonomous driving as it describes driving scenes in a comprehensive manner. 
This task aims to label each voxel in the scene by taking one or more types of sensors as input. 
Two most used sensors are LiDAR and the camera. 
Although LiDAR-based methods perform remarkably well in 3D perception tasks~\cite{cheng20212s3net, liong2020amvnet, tang2020searching, ye2023lidarmultinet, ye2021drinet++, lang2019pointpillars, zhou2018voxelnet, chen2017multi, yan2021JS3CNet, lmscnet, aicnet, 3DSketch, cao2022monoscene}, they possess limitations under bad weather conditions and in detecting distant objects; 
Thus, camera-based approaches have garnered increasing attention~\cite{wei2023surroundocc,yu2023flashocc,li2023fb}. 
Pioneer works in 3D semantic occupancy prediction task adopt dense grid-based representation as a straightforward mean to derive occupancy~\cite{li2023voxformer,wei2023surroundocc,3Chen20193DSS}, then subsequent works turn to sparse object-centric representation~\cite{tang2024sparseocc,lu2023octreeocc,huang2024gaussian} as a solution to the innate problem of redundancy for dense representations. 

\textbf{Grid-based scene representations.}
Plane representations have emerged as competitive representations in scene perception tasks for autonomous driving. 
BEVFormer~\cite{li2022bevformer} is the initiative work of the kind~\cite{huang2021bevdet, li2022bevdepth, philion2020lss, liang2022bevfusion, liu2023bevfusion} that utilizes only camera input and performs comparably well with LiDAR-based methods in detection and segmentation tasks. 
It converts the image feature to the bird's-eye-view (BEV) feature as a unified scene representation, since the information is most diverse at this point of view.
The BEV feature is then used for downstream tasks.
However, the BEV feature is not suitable for 3D occupancy construction as it causes height information to be lost~\cite{wei2023surroundocc}. 
As a generalization of BEV space, TPVFormer~\cite{ huang2023tri} proposes tri-perspective view representation to include also the height information, thus making it more suitable for 3D scenes. 
Another research direction~\cite{wei2023surroundocc, li2023voxformer} adopts voxel-based representation as a more 3D-specific and fine-grained approach, making it favorable for 3D volumetric semantic prediction. 
Nevertheless, these methods utilize dense grid-based representation, which describes each voxel equally regardless of the spatial sparsity of the environment, thus resulting in intrinsic redundancy.

\textbf{Object-centric scene representations.}
To eliminate spatial redundancy inherent in dense representations, many recent works adopt sparse representation~\cite{tang2024sparseocc,lu2023octreeocc,huang2024gaussian}. 
One line of work divides dense grids into partitions where objects present and omits the regions foreseen as empty~\cite{lu2023octreeocc, tang2024sparseocc}. 
However, non-empty regions might be mistakenly classified as unoccupied and eliminated completely from the whole subsequent process. 
Another line of work leverages point representation ~\cite{shi2024occupancysetpoints, wang2024opus} by sampling points within the scene range as queries in the succeeding refinement process; 
Nevertheless, a point has a limited range of depiction as it has no spatial extent. 
An alternative approach, GaussianFormer~\cite{huang2024gaussian}, adopts 3D semantic Gaussian representation, where probability spreads around a mean, allowing more utilization. However, spatial redundancy persists due to no regulation for the Gaussians to not represent emptiness. 

\section{Proposed Approach}
\label{sec: approach}
In this section, we present our method of probabilistic Gaussian superposition for efficient 3D semantic occupancy prediction.
We first review the original 3D semantic Gaussian representation~\cite{huang2024gaussian} and its limitations (Sec.~\ref{subsec: old gaussian}).
We then introduce our probabilistic Gaussian modeling and how we derive geometry and semantics predictions based on the multiplication theorem of probability and Gaussian mixture model (Sec.~\ref{subsec: prob modeling}).
Finally, we detail the distribution-based initialization module to effectively initialize probabilistic Gaussians around the occupied area (Sec.~\ref{subsec: initialization}).

\subsection{3D Semantic Gaussian Representation}
\label{subsec: old gaussian}
Vision-centric 3D semantic occupancy prediction~\cite{cao2022monoscene,huang2023tri} aims to obtain the fine-grained geometry and semantics of the 3D scene.
To formulate, the target is to predict voxel-level semantic segmentation result $\mathbf{O}\in \mathcal{C}^{X\times Y\times Z}$ given input images $\mathcal{I}=\{\mathbf{I}_i\}_{i=1}^{N}$, where $\mathcal{C}$, $\{X, Y, Z\}$, $N$ represent the set of predefined classes, the spatial resolution of occupancy and the number of input views, respectively.

To achieve this, 3D semantic Gaussian representation employs a set of $P$ Gaussian primitives $\mathcal{G}=\{\mathbf{G}_i\}_{i=1}^{P}$, with each $\mathbf{G}_i$ describing a local region with its mean $\mathbf{m}_i$, scale $\mathbf{s}_i$, rotation $\mathbf{r}_i$, opacity $a_i$ and semantics $\mathbf{c}_i$.
GaussianFormer interprets these primitives as local semantic Gaussian distributions which contribute to the overall occupancy prediction through additive aggregation:
\vspace{-2mm}
\begin{equation}
    \hat{\mathbf{o}}(\mathbf{x}; \mathcal{G})=\sum_{i=1}^{P}\mathbf{g}_i(\mathbf{x}; \mathbf{m}_i,\mathbf{s}_i,\mathbf{r}_i,a_i,\mathbf{c}_i),
    \label{eq: weighted summation}
    \vspace{-2mm}
\end{equation}
where $\mathbf{g}_i(\mathbf{x};\cdot)$ denotes the contribution of the $i$th semantic Gaussian to $\hat{\mathbf{o}}(\mathbf{x}; \mathcal{G})$ which is the overall occupancy prediction at location $\mathbf{x}$.
The contribution $\mathbf{g}$ is further calculated as the corresponding semantic Gaussian distribution evaluated at location $\mathbf{x}$:
\begin{equation}
    \mathbf{g}(\mathbf{x}; \mathbf{G}) = a\cdot{\rm{exp}}\big(-\frac{1}{2}(\mathbf{x}-\mathbf{m})^{\rm T} \mathbf{\Sigma}^{-1} (\mathbf{x}-\mathbf{m})\big)\mathbf{c},
    \label{eq: gaussian dist}
\end{equation}
\begin{equation}
    \mathbf{\Sigma} = \mathbf{R}\mathbf{S}\mathbf{S}^T\mathbf{R}^T, \quad \mathbf{S} = {\rm{diag}}(\mathbf{s}), \quad \mathbf{R} = {\rm{q2r}}(\mathbf{r}),
\end{equation}
where $\mathbf{\Sigma}$, $\mathbf{R}$, $\mathbf{S}$ represent the covariance matrix, the rotation matrix constructed from the quaternion $\mathbf{r}$ with function ${\rm q2r}(\cdot)$, and the diagonal scale matrix from function ${\rm diag}(\cdot)$.

Although the number of Gaussians is reduced compared with the number of dense voxels thanks to the deformable nature of Gaussian distributions as in Eq.~(\ref{eq: gaussian dist}), several limitations still persist in the 3D semantic Gaussian representation.
First of all, it models both the occupied and unoccupied regions in the same way using the semantic property $\mathbf{c}$, resulting in most Gaussians being classified as empty given the huge proportion of empty space in outdoor scenarios.
Secondly, the semantic Gaussian representation encourages Gaussians to overlap, because the aggregation process in Eq.~(\ref{eq: weighted summation}) independently sums up the contribution of each Gaussian, resulting in unbounded occupancy prediction $\hat{\mathbf{o}}$.
For optimization, the model would learn to allocate more Gaussians to describe the same region due to the unbounded nature of $\hat{\mathbf{o}}$, aggravating the overlap between Gaussians.
These limitations stem from the current interpretation of Gaussians and obstruct the efficiency and effectiveness of the 3D semantic Gaussian representation.
Our method approaches Gaussian-based object-centric representation from a probabilistic perspective, serving as a fundamental solution to these issues, as shown by Figure~\ref{fig:motivation}.

\subsection{Probabilistic Gaussian Superposition}
\label{subsec: prob modeling}
We propose the probabilistic Gaussian superposition as an efficient and effective 3D scene representation.
As shown in Figure~\ref{fig:pipeline}, we decompose the 3D modeling target into geometry and semantics predictions, and adopt the multiplication theorem of probability and the Gaussian mixture model to address them from a probabilistic perspective, respectively.

\textbf{Geometry prediction.}
To restrict Gaussians to represent only occupied regions for geometry prediction, we interpret the Gaussian primitives $\mathcal{G}=\{\mathbf{G}_i\}_{i=1}^{P}$ as the probability of their surrounding space being occupied.
To elaborate, we assign a probability value of 100\% at the centers of Gaussians, which decays exponentially with respect to the distance from the centers $\mathbf{m}$:
\begin{equation}
    \alpha(\mathbf{x};\mathbf{G}) = {\rm{exp}}\big(-\frac{1}{2}(\mathbf{x}-\mathbf{m})^{\rm T} \mathbf{\Sigma}^{-1} (\mathbf{x}-\mathbf{m})\big),
    \label{eq: single prob}
\end{equation}
where $\alpha(\mathbf{x};\mathbf{G})$ denotes the probability of the point $\mathbf{x}$ being occupied induced by Gaussian $\mathbf{G}$.
Eq.~(\ref{eq: single prob}) assigns a high probability of occupancy when the point $\mathbf{x}$ is close to the center of Gaussian $\mathbf{G}$, which prevents any Gaussian from describing empty area.
To further derive the overall probability of occupancy, we assume that the probabilities of a point being occupied by different Gaussians are mutually independent, and thus we can aggregate them according to the multiplication theorem of probability:
\vspace{-1mm}
\begin{equation}
    \alpha(\mathbf{x}) = 1 - \prod_{i=1}^{P}\big(1 - \alpha(\mathbf{x};\mathbf{G}_i)\big),
    \label{eq: multi prob}
    \vspace{-1mm}
\end{equation}
where $\alpha(\mathbf{x})$ represents the overall probability of occupancy at point $\mathbf{x}$. 
In addition to achieving object-centric properties, Eq.~(\ref{eq: multi prob}) also avoids unnecessary overlapping between Gaussians because $\alpha(\mathbf{x}) \ge \alpha(\mathbf{x};\mathbf{G}_i)$ holds for any Gaussian $\mathbf{G}_i$.
This implies that point $\mathbf{x}$ would be predicted occupied if it is close enough to any single Gaussian.

\textbf{Semantics prediction.}
In addition to object-centric anti-overlapping geometry modeling, we still need to achieve the same goals for semantics prediction.
We first remove the channel that represents the empty class from the semantic properties $\mathbf{c}$ of Gaussians since it has been accounted for in geometry prediction.
Then we interpret the set of Gaussians $\mathcal{G}$ as a Gaussian mixture model, where semantics prediction could be formulated as calculating the expectation of semantics given the probabilistic Gaussian mixture model.
Specifically, we take the original opacity properties $a$ as the prior distribution of Gaussians, which is $l^1$-normalized.
Furthermore, we adopt the Gaussian probabilistic distribution parameterized by mean $\mathbf{m}$, scale $\mathbf{s}$ and rotation $\mathbf{r}$ as the conditional probability.
Then we normalize the original semantics properties $\mathbf{c}$ with softmax to ensure the boundedness of predicted semantics.
Finally, we calculate the expectation $\mathbf{e}(\mathbf{x};\mathcal{G})$ as:
\vspace{-1mm}
\begin{equation}
\begin{aligned}
    \mathbf{e}(\mathbf{x};\mathcal{G}) &= \sum_{i=1}^{P} p(\mathbf{G}_i|\mathbf{x})\Tilde{\mathbf{c}}_i 
    = \frac{\sum_{i=1}^{P}p(\mathbf{x}|\mathbf{G}_i)a_i\Tilde{\mathbf{c}}_i}{\sum_{j=1}^{P}p(\mathbf{x}|\mathbf{G}_j)a_j},
\end{aligned}
\label{eq: gmm}
\end{equation}
\begin{small}
\begin{equation}
    p(\mathbf{x}|\mathbf{G}_i) = \frac{1}{(2\pi)^{\frac{3}{2}}|\mathbf{\Sigma}|^{\frac{1}{2}}}{\rm{exp}}\big(-\frac{1}{2}(\mathbf{x}-\mathbf{m})^{\rm T} \mathbf{\Sigma}^{-1} (\mathbf{x}-\mathbf{m})\big),
\end{equation}
\end{small}where $p(\mathbf{G}_i|\mathbf{x})$, $p(\mathbf{x}|\mathbf{G}_i)$ and $\Tilde{\mathbf{c}}_i$ denote the posterior probability of point $\mathbf{x}$ belonging to the $i$th Gaussian distribution, the conditional probability of point $\mathbf{x}$ given the $i$th Gaussian distribution, and the softmax-normalized semantic properties, respectively. 
Compared with Eq.~(\ref{eq: weighted summation})(\ref{eq: gaussian dist}), the gaussian mixture model in Eq.~(\ref{eq: gmm}) normalizes the semantic properties and the contributions from different Gaussians, thus preventing unnecessary overlapping between Gaussians and producing normalized class probabilities directly.

Given the geometry and semantics predictions, we take a simple step forward to combine them to generate the final semantic occupancy prediction:
\begin{equation}
    \hat{\mathbf{o}}(\mathbf{x};\mathcal{G}) = [1-\alpha(\mathbf{x}); \alpha(\mathbf{x})\cdot\mathbf{e}(\mathbf{x};\mathcal{G})],
    \label{eq: final occ}
\end{equation}
where we use the geometry probability $\alpha(\mathbf{x})$ to weight the semantic predictions, and directly take $1 - \alpha(\mathbf{x})$ as the probability of the empty class.

\begin{figure}[t]
\centering
\includegraphics[width=0.95\linewidth]{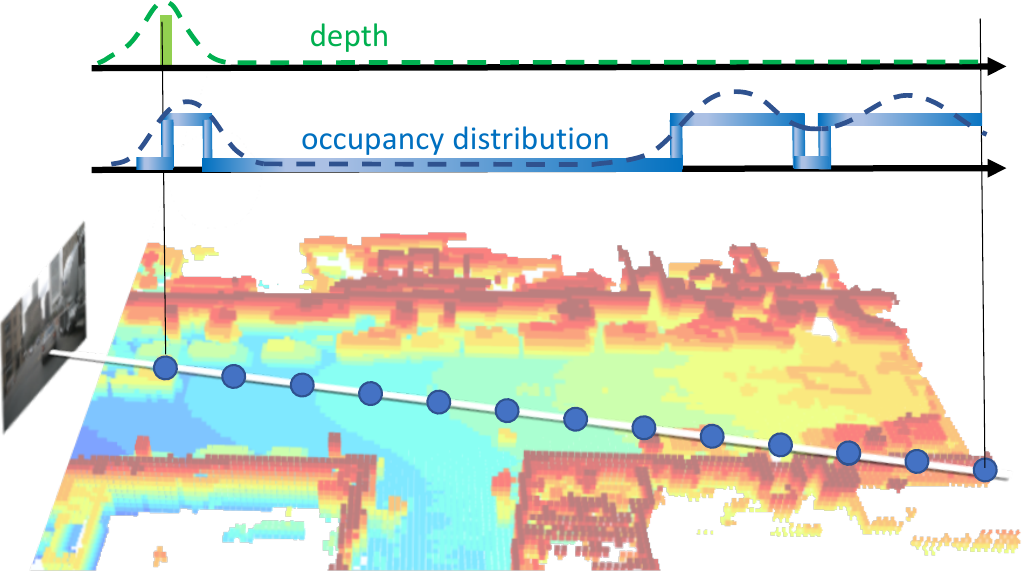}
\vspace{-2mm}
\caption{\textbf{Distribution-based initialization.}
Our initialization scheme learns pixel-aligned occupancy distributions from occupancy annotation, while the depth-based counterpart only captures the surfaces of objects and relies on LiDAR supervision.
}
\label{fig:initialization}
\vspace{-6mm}
\end{figure}

\begin{table*}[t] %
    \caption{\textbf{Surround view 3D semantic occupancy prediction results on nuScenes.} 
    * means supervised by dense occupancy annotations as opposed to original LiDAR segmentation labels.
    Ch. denotes the channel dimension of our model.
    Our method achieves state-of-the-art performance compared with other methods.}
    \small
    \setlength{\tabcolsep}{0.005\linewidth}  
    \vspace{-3mm}  
    \renewcommand\arraystretch{1.05}
    \centering
    \resizebox{\textwidth}{!}{
    \begin{tabular}{l|c c | c c c c c c c c c c c c c c c c}
        \toprule
        Method
        & IoU
        & mIoU
        & \rotatebox{90}{\textcolor{nbarrier}{$\blacksquare$} barrier}
        & \rotatebox{90}{\textcolor{nbicycle}{$\blacksquare$} bicycle}
        & \rotatebox{90}{\textcolor{nbus}{$\blacksquare$} bus}
        & \rotatebox{90}{\textcolor{ncar}{$\blacksquare$} car}
        & \rotatebox{90}{\textcolor{nconstruct}{$\blacksquare$} const. veh.}
        & \rotatebox{90}{\textcolor{nmotor}{$\blacksquare$} motorcycle}
        & \rotatebox{90}{\textcolor{npedestrian}{$\blacksquare$} pedestrian}
        & \rotatebox{90}{\textcolor{ntraffic}{$\blacksquare$} traffic cone}
        & \rotatebox{90}{\textcolor{ntrailer}{$\blacksquare$} trailer}
        & \rotatebox{90}{\textcolor{ntruck}{$\blacksquare$} truck}
        & \rotatebox{90}{\textcolor{ndriveable}{$\blacksquare$} drive. suf.}
        & \rotatebox{90}{\textcolor{nother}{$\blacksquare$} other flat}
        & \rotatebox{90}{\textcolor{nsidewalk}{$\blacksquare$} sidewalk}
        & \rotatebox{90}{\textcolor{nterrain}{$\blacksquare$} terrain}
        & \rotatebox{90}{\textcolor{nmanmade}{$\blacksquare$} manmade}
        & \rotatebox{90}{\textcolor{nvegetation}{$\blacksquare$} vegetation}
        \\
        \midrule
        MonoScene~\cite{cao2022monoscene} & 23.96 & 7.31 & 4.03 &	0.35& 8.00& 8.04&	2.90& 0.28& 1.16&	0.67&	4.01& 4.35&	27.72&	5.20& 15.13&	11.29&	9.03&	14.86 \\
        
        Atlas~\cite{murez2020atlas} & 28.66 & 15.00 & 10.64&	5.68&	19.66& 24.94& 8.90&	8.84&	6.47& 3.28&	10.42&	16.21&	34.86&	15.46&	21.89&	20.95&	11.21&	20.54 \\
        
        BEVFormer~\cite{li2022bevformer} & 30.50 & 16.75 & 14.22 &	6.58 & 23.46 & 28.28& 8.66 &10.77& 6.64& 4.05& 11.20&	17.78 & 37.28 & 18.00 & 22.88 & 22.17 & {13.80} &	\textbf{22.21}\\
        
        TPVFormer~\cite{huang2023tri} & 11.51 & 11.66 & 16.14&	7.17& 22.63	& 17.13 & 8.83 & 11.39 & 10.46 & 8.23&	9.43 & 17.02 & 8.07 & 13.64 & 13.85 & 10.34 & 4.90 & 7.37\\
        
        TPVFormer*~\cite{huang2023tri}  & {30.86} & 17.10 & 15.96&	 5.31& 23.86	& 27.32 & 9.79 & 8.74 & 7.09 & 5.20& 10.97 & 19.22 & {38.87} & {21.25} & {24.26} & {23.15} & 11.73 & 20.81\\

        OccFormer~\cite{zhang2023occformer} & {31.39} & {19.03} & {18.65} & {10.41} & {23.92} & {30.29} & {10.31} & {14.19} & {13.59} & {10.13} & {12.49} & {20.77} & {38.78} & 19.79 & 24.19 & 22.21 & {13.48} & {21.35}\\
        
        SurroundOcc~\cite{wei2023surroundocc} & {31.49} & {20.30}  & {20.59} & {11.68} & {28.06} & \textbf{30.86} & {10.70} & {15.14} & \textbf{14.09} & \textbf{12.06} & \textbf{14.38} & {22.26} & 37.29 & {23.70} & {24.49} & {22.77} & \textbf{14.89} & {21.86}  \\

        GaussianFormer~\cite{huang2024gaussian} & 29.83 & {19.10} & {19.52} & {11.26} & {26.11} & {29.78} & {10.47} & {13.83} & {12.58} & {8.67} & {12.74} & {21.57} & {39.63} & {23.28} & {24.46} & {22.99} & 9.59 & 19.12 \\

        \midrule
        \textbf{Ours} (Ch. = 128) & 30.56 & 20.02 & 20.15 & 12.99 & 27.61 & 30.23 & \textbf{11.19} & 15.31 & 12.64 & 9.63 & 13.31 & 22.26 & 39.68 & 23.47 & 25.62 & 23.20 & 12.25 & 20.73 \\
        
        \textbf{Ours} (Ch. = 192) & \textbf{31.74} & \textbf{20.82} & \textbf{21.39} & \textbf{13.44} & \textbf{28.49} & 30.82 & 10.92 & \textbf{15.84} & 13.55 & 10.53 & 14.04 & \textbf{22.92} & \textbf{40.61} & \textbf{24.36} & \textbf{26.08} & \textbf{24.27} & 13.83 & 21.98  \\
        
        \bottomrule
    \end{tabular}}
    \label{tab: nuscenes results}
    \vspace{-5mm}
\end{table*}

\subsection{Distribution-Based Initialization}
\label{subsec: initialization}
Previous 3D semantic Gaussian representation adopts a learnable initialization strategy, which randomly initializes the properties of Gaussians at the beginning of training, and optimizes this initialization in a data-driven way.
This strategy enables the model to learn a prior distribution of occupancy of the whole dataset, which relies on the subsequent refinement of the network to adapt to the distribution of each individual sample.
However, the local receptive field of Gaussians limits their mobility, which hinders each Gaussian from learning the path to the correct position in subsequent refinement.
And this issue is even more severe for our probabilistic Gaussian superposition where Gaussians are supposed to model only occupied regions.

To remedy this issue, we propose a distribution-based initialization module which provides both more accurate and holistic sample-specific initialization for Gaussians, as shown by Figure~\ref{fig:initialization}.
We supervise the image features from a 2D backbone with the pixel-aligned occupancy distribution derived from the occupancy annotations.
To elaborate, we first determine the origin $\mathbf{b}$ and direction $\mathbf{d}$ of the ray corresponding to each image feature with the camera calibration data.
We then sample $R$ reference points at equal intervals in a fixed depth range along this ray.
For each of these reference points, we query the ground truth occupancy $\mathbf{O}$ at the corresponding location to obtain the binary labels $\mathbf{l}=\{l_i\}_{i=1}^{R}$ indicating whether a reference point is occupied or not.
Then we use $\mathbf{l}=\{l_i\}_{i=1}^{R}$ as supervision to optimize our initialization module, which consists of an image backbone ${\rm B}$ and a distribution predictor ${\rm M}$.
The distribution predictor ${\rm M}$ directly decodes image features into occupancy distributions $\hat{\mathbf{l}}$ along corresponding rays, which are matched against $\mathbf{l}$ using binary cross entropy loss:
\begin{equation}
    loss_{init} = {\rm BCE}\big(\hat{\mathbf{l}}, \mathbf{l}\big) = {\rm BCE}\big({\rm M}({\rm B}(\mathcal{I})), \mathbf{l}\big).
\end{equation}
Different from previous initialization schemes~\cite{li2023voxformer,li2022bevdepth,huang2021bevdet} that predict the depth values with LiDAR supervision, our method learns the holistic occupancy distribution rather than only visible surfaces of the scene, and does not require any additional modality as supervision.

Overall, our distribution-based initialization module initializes the Gaussians, which are subsequently sent into B blocks of attention-based architecture as in GaussianFormer~\cite{huang2024gaussian}. 
Each block consists of self-encoding, image cross-attention, and refinement module, where probabilistic Gaussian properties steadily improve, then the resulting Gaussians are aggregated by our new method that encourages higher utilization of Gaussians.

\section{Experiments}
\label{sec: experiments}

\begin{table*}[t] %
    \centering
    \caption{\textbf{Monocular 3D semantic occupancy prediction results on SSCBench-KITTI-360.}
    Our method achieves state-of-the-art performance compared with other methods, surpassing GaussianFormer~\cite{huang2024gaussian} by a clear margin.
    }
    \label{tab:kittiseg}
    \vspace{-3mm}
    \setlength{\tabcolsep}{0.005\linewidth}   
    \renewcommand\arraystretch{1.05}
    \resizebox{1\linewidth}{!}{
    \begin{tabular}{l|c|c|c| c c c c c c c c c c c c c c c c c c}
    \toprule
    Method 
    & {\rotatebox{90}{Input}}  
    & IoU
    & mIoU
    &\rotatebox{90}{\textcolor{carcolor}{$\blacksquare$} car}
    &\rotatebox{90}{\textcolor{bicyclecolor}{$\blacksquare$} {bicycle}}
    &\rotatebox{90}{\textcolor{motorcyclecolor}{$\blacksquare$} {motorcycle}}
    &\rotatebox{90}{\textcolor{truckcolor}{$\blacksquare$} {truck}}
    &\rotatebox{90}{\textcolor{othervehiclecolor}{$\blacksquare$} {other-veh.}}
    &\rotatebox{90}{\textcolor{personcolor}{$\blacksquare$} {person}}
    &\rotatebox{90}{\textcolor{roadcolor}{$\blacksquare$} {road}}  
    &\rotatebox{90}{\textcolor{parkingcolor}{$\blacksquare$} {parking}}
    &\rotatebox{90}{\textcolor{sidewalkcolor}{$\blacksquare$} {sidewalk}}
    &\rotatebox{90}{\textcolor{othergroundcolor}{$\blacksquare$} {other-grnd}}
    &\rotatebox{90}{\textcolor{buildingcolor}{$\blacksquare$} {building}}
    &\rotatebox{90}{\textcolor{fencecolor}{$\blacksquare$} {fence}}
    &\rotatebox{90}{\textcolor{vegetationcolor}{$\blacksquare$} {vegetation}}
    &\rotatebox{90}{\textcolor{terraincolor}{$\blacksquare$} {terrain}}
    &\rotatebox{90}{\textcolor{polecolor}{$\blacksquare$} {pole}}
    &\rotatebox{90}{\textcolor{trafficsigncolor}{$\blacksquare$} {traf.-sign}}
    &\rotatebox{90}{\textcolor{other-struct.color}{$\blacksquare$} {other-struct.}}
    &\rotatebox{90}{\textcolor{other-objectcolor}{$\blacksquare$} {other-object}}
     
    \\\midrule

    LMSCNet~\cite{lmscnet} & L & {47.53} & {13.65} & {20.91} & {0} & {0} & {0.26} & {0} & {0} & {62.95} & {13.51} & {33.51} & {0.2} & {43.67} & {0.33} & {40.01} & {26.80} & {0} & {0} & {3.63} & {0}
        
    \\ SSCNet~\cite{song2017semantic} & L & {53.58} & {16.95} & {31.95} & {0} & {0.17} & {10.29} & {0.58} & {0.07} & {65.7} & {17.33} & {41.24} & {3.22} & {44.41} & {6.77} & {43.72} & {28.87} & {0.78} & {0.75} & {8.60} & {0.67} 
    
    \\\midrule MonoScene~\cite{cao2022monoscene} & C & {37.87} & {12.31} & {19.34} & {0.43} & {0.58} & {8.02} & {2.03} & {0.86} & {48.35} & {11.38} & {28.13} & {3.22} & {32.89} & {3.53} & {26.15} & {16.75} & {6.92} & {5.67} & {4.20} & {3.09}
    
    \\ Voxformer~\cite{li2023voxformer} & C & {38.76} & {11.91} & {17.84} & {1.16} & {0.89}& {4.56} & {2.06}  & {1.63} & {47.01} & {9.67} & {27.21} & {2.89} & {31.18} & {4.97} & {28.99} & {14.69} & {6.51} & {6.92} & {3.79} & {2.43}
    
    \\ TPVFormer~\cite{huang2023tri} & C & {40.22} & {13.64} & {21.56} & {1.09} & {1.37} & {8.06} & {2.57} & {2.38} & {52.99} & {11.99} & {31.07} & {3.78} & {34.83} & {4.80} & {30.08} & {17.51} & {7.46} & {5.86} & {5.48} & {2.70} 
    
    \\ OccFormer~\cite{zhang2023occformer} & C & \textbf{40.27} & {13.81} & \textbf{22.58} & {0.66} & {0.26} & {9.89} & {3.82} & {2.77} & \textbf{54.30} & \textbf{13.44} & {31.53} & {3.55} & \textbf{36.42} & {4.80} & \textbf{31.00} & \textbf{19.51} & \textbf{7.77} & \textbf{8.51} & {6.95} & {4.60}

    \\ GaussianFormer~\cite{huang2024gaussian} & C & 35.38 & {12.92} & 18.93 & {1.02} & \textbf{4.62} & \textbf{18.07} & \textbf{7.59} & \textbf{3.35} & 45.47 & 10.89 & 25.03 & \textbf{5.32} & 28.44 & \textbf{5.68} & {29.54} & 8.62 & 2.99 & 2.32 & \textbf{9.51} & \textbf{5.14}

    \\\midrule \textbf{Ours} & C & 38.37 & \textbf{13.90} & 21.08 & \textbf{2.55} & 4.21 & 12.41 & 5.73 & 1.59 & 54.12 & 11.04 & \textbf{32.31} & 3.34 & 32.01 & 4.98 & 28.94 & 17.33 & 3.57 & 5.48 & 5.88 & 3.54
 
\\\bottomrule
\end{tabular}
}
\vspace{-5mm}
\end{table*}

\subsection{Datasets and Metrics}

\textbf{The nuScenes dataset}~\cite{caesar2020nuscenes} provides 1000 scenes of surround view driving scenes in Boston and Singapore. 
The official division is 700/150/150 scenes for training, validation, and testing, respectively. 
Each scene is 20 seconds long and fully annotated at 2Hz with ground truth from 5 radars, 6 cameras, one LiDAR, and one IMU. 
We employ 3D semantic occupancy annotations from SurroundOcc~\cite{wei2023surroundocc} for supervision and evaluation. 
The ranges of the occupancy annotations in the x, y, and z axes in meters are [-50, 50], [-50, 50], and [-5, 3], respectively, where each voxel has a side length of 0.5 meters and is labeled as one of the 18 possible classes (16 semantics, 1 empty, and 1 noise class).

\textbf{The KITTI-360 dataset}~\cite{Liao2022kitti360} consists of over 320k images in suburban area with rich 360 degree sensory ground truth, consisting of 2 perspective cameras, 2 fisheye cameras, a Velodyne LiDAR, and a laser scanner, where we use the images from the left camera of the ego car as input to our model. 
For 3D semantic occupancy prediction, we adopt the annotations from SSCBench-KITTI-360~\cite{li2023sscbench}. 
The official split is 7/1/1 sequences with 8487/1812/2566 key frames for training, validation, and testing, respectively. 
The voxel grid area covers 51.2$\times$51.2$\times$6.4 $m^2$ in front of the ego car with resolution of 256$\times$256$\times$32. 
Each voxel is classified as one of the 19 classes (18 semantics and 1 empty).

\textbf{The evaluation metrics} are in accordance with common practice~\cite{cao2022monoscene}, namely mean Intersection-over-Union (mIoU) and Intersection-over-Union (IoU):
\begin{equation}
\mathbf{mIoU} = \frac{1}{|\mathcal{C}'|}\sum_{i\in \mathcal{C}'}{\frac{TP_i}{TP_i+FP_i+FN_i}},
\end{equation}
\begin{equation}
\mathbf{IoU} = \frac{TP_{\neq c_0}}{TP_{\neq c_0}+FP_{\neq c_0}+FN_{\neq c_0}},
\end{equation}
Where $\mathcal{C}'$, $c_0$, TP, FP, and FN represent the non-empty classes, the empty class, and the number of true positive, false positive, and false negative predictions, respectively.

\subsection{Implementation Details}
The input images are at resolutions of 900$\times$1600 for nuScenes~\cite{caesar2020nuscenes} and 376x1408 for KITTI-360~\cite{Liao2022kitti360} with random flipping and photometric distortion augmentations. 
We use the same checkpoints for our image backbone as used in GaussianFormer~\cite{huang2024gaussian}, i.e. ResNet101-DCN~\cite{he2016resnet} with FCOS3D checkpoint~\cite{wang2021fcos3d} for nuScenes, and ResNet50~\cite{he2016resnet} pretrained on ImageNet~\cite{deng2009imagenet} for KITTI-360. 
The numbers of Gaussians are set to 12800 and 38400 in our main results for nuScenes and KITTI-360, respectively. 
We train our model using AdamW~\cite{loshchilov2017adamw} with weight decay of 0.01, and maximum learning rate of $2\times 10^{-4}$, which decays with a cosine annealing schedule. 
We train our model for 20 epochs on nuScenes with a batch of 8 and 30 epochs on KITTI-360 with a batch size of 4, respectively.

\subsection{Main Results}

\textbf{Surround-view 3D semantic occupancy prediction.}
We report the performance of our GaussianFormer-2 in Table~\ref{tab: nuscenes results}.
Our approach achieves state-of-the-art performance compared with other methods.
Specifically, GaussianFormer-2 surpasses methods based on dense grid representation in classes such as bicycle and motorcycle, proving the flexibility of the proposed probabilistic Gaussian superposition in modeling small objects.
Furthermore, our method outperforms GaussianFormer~\cite{huang2024gaussian} with a clear margin and significantly fewer Gaussians (12800 v.s. 144000), which validates the efficiency and effectiveness of our method.

\textbf{Monocular 3D semantic occupancy prediction.}
We report the results for monocular 3D semantic occupancy prediction on SSCBench-KITTI-360~\cite{li2023sscbench} in Table~\ref{tab:kittiseg}.
Our method achieves state-of-the-art performance, surpassing the original GaussianFormer in mIoU by 7.6\%.
To elaborate, we observe significant improvement in mIoU of classes such as road, sidewalk and building compared with GaussianFormer, showing the superiority of probabilistic Gaussian superposition in modeling background staff.

\begin{table}[t]
    \centering
    \caption{
    \textbf{Ablation on the number of Gaussians.}
    The latency and memory are tested on an NVIDIA 4090 GPU with batch size one during inference, in accordance with GaussianFormer~\cite{huang2024gaussian}.
    We report the latency of the initialization module in parentheses.
    Our method achieves better performance-efficiency trade-off.
    }
    \vspace{-2mm}
    \setlength{\tabcolsep}{0.02\linewidth}
    \resizebox{1\linewidth}{!}{
    \begin{tabular}{c|ccc|cc}
        \toprule
        Method & \makecell{Number of\\ Gaussians} & \makecell{Latency\\(ms)} & \makecell{Memory\\(MB)} & mIoU & IoU \\
        \midrule
        \multirow{2}{*}{\makecell{Gaussian-\\Former}} & 25600  & \textbf{227} & {4850}  & 16.00  & 28.72 \\
        & 144000 & {372} & 6229  & {19.10}  & {29.83}  \\
        \midrule
        \multirow{3}{*}{{Ours}} & 6400  & \underline{313} (142) & \textbf{3026} & 19.87 & \underline{30.37} \\
                                       & 12800 & 323 (143) & \underline{3041} & \underline{19.94} & \underline{30.37} \\
                                       & 25600 & 357 (147) & 3063 & \textbf{20.33} & \textbf{31.04} \\
        \bottomrule
    \end{tabular}}
    \vspace{-4mm}
    \label{tab:number of gaussians}
\end{table}

\begin{table}[t]
    \centering
    \caption{
    \textbf{Ablation on the components of GaussianFormer-2.}
    We set the number of Gaussians to 25600 for these experiments.
    Depth means using depth as supervision in the initialization module instead of occupancy distribution.
    Pointcloud represents using ground truth LiDAR scan for initialization.
    }
    \vspace{-2mm}
    \setlength{\tabcolsep}{0.055\linewidth}
    \resizebox{1\linewidth}{!}{
    \begin{tabular}{cc|cc}
        \toprule
        \makecell{Probabilistic \\ Modeling} & \makecell{Gaussian \\ Initialization}  & mIoU & IoU \\
        \midrule
        &   & 16.00 & 28.72 \\
        \checkmark &               & 19.61 & 30.61 \\
        \checkmark & Depth         & 19.97 & 30.87 \\
        \checkmark & Pointcloud    & \textbf{21.17} & \textbf{34.91} \\
        \checkmark & Distribution  & \underline{20.32} & \underline{31.04} \\
        \bottomrule
    \end{tabular}}
    \vspace{-7mm}
    \label{tab:components}
\end{table}

\begin{figure*}[t]
\centering
\includegraphics[width=0.95\linewidth]{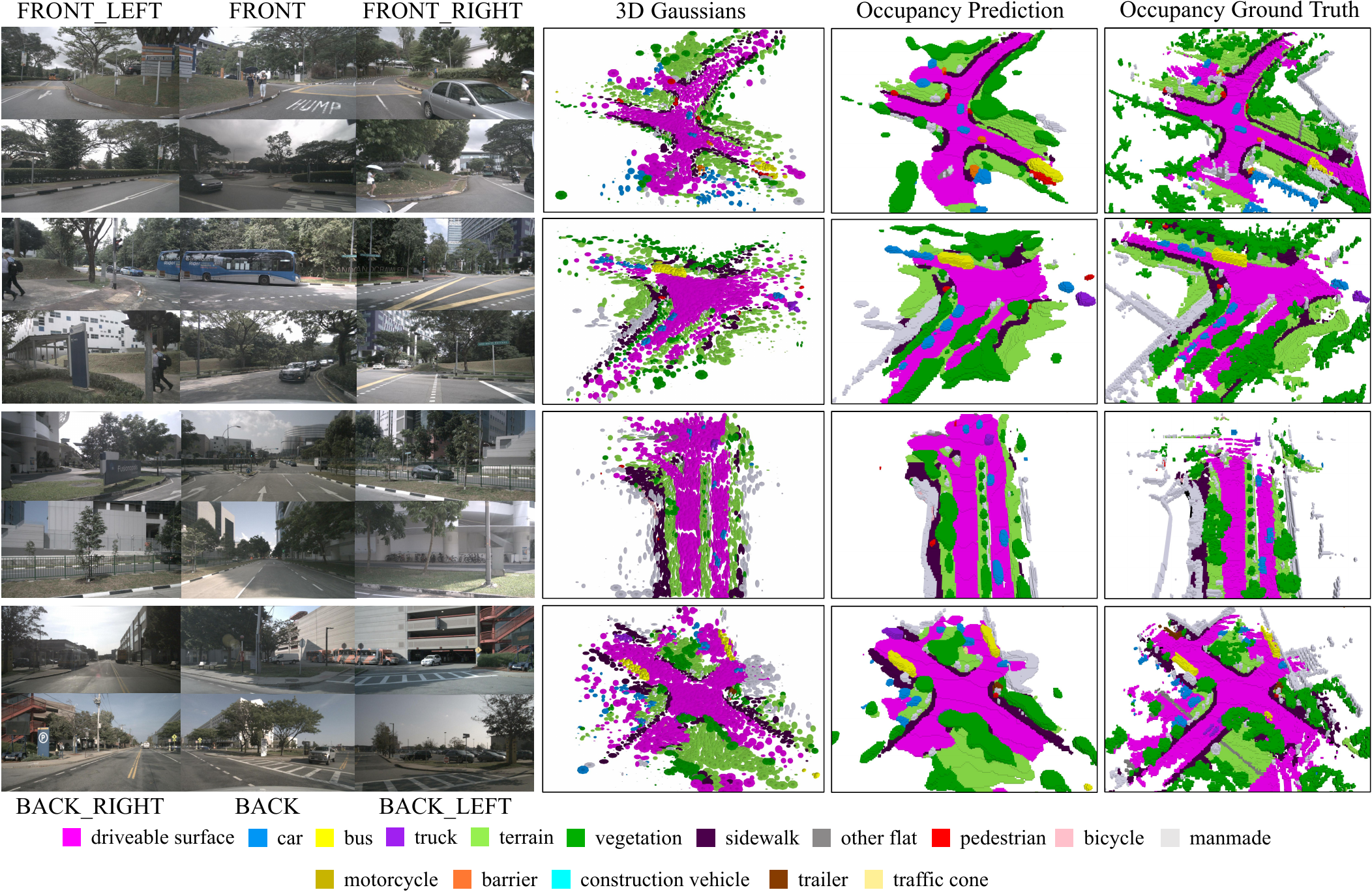}
\vspace{-2mm}
\caption{\textbf{Gaussian and occupancy visualizations on nuScenes.}
Our model is able to predict both comprehensive and realistic 3D Gaussians and occupancy.
}
\label{fig:main}
\vspace{-6mm}
\end{figure*}

\begin{figure*}[t]
\centering
\includegraphics[width=0.95\linewidth]{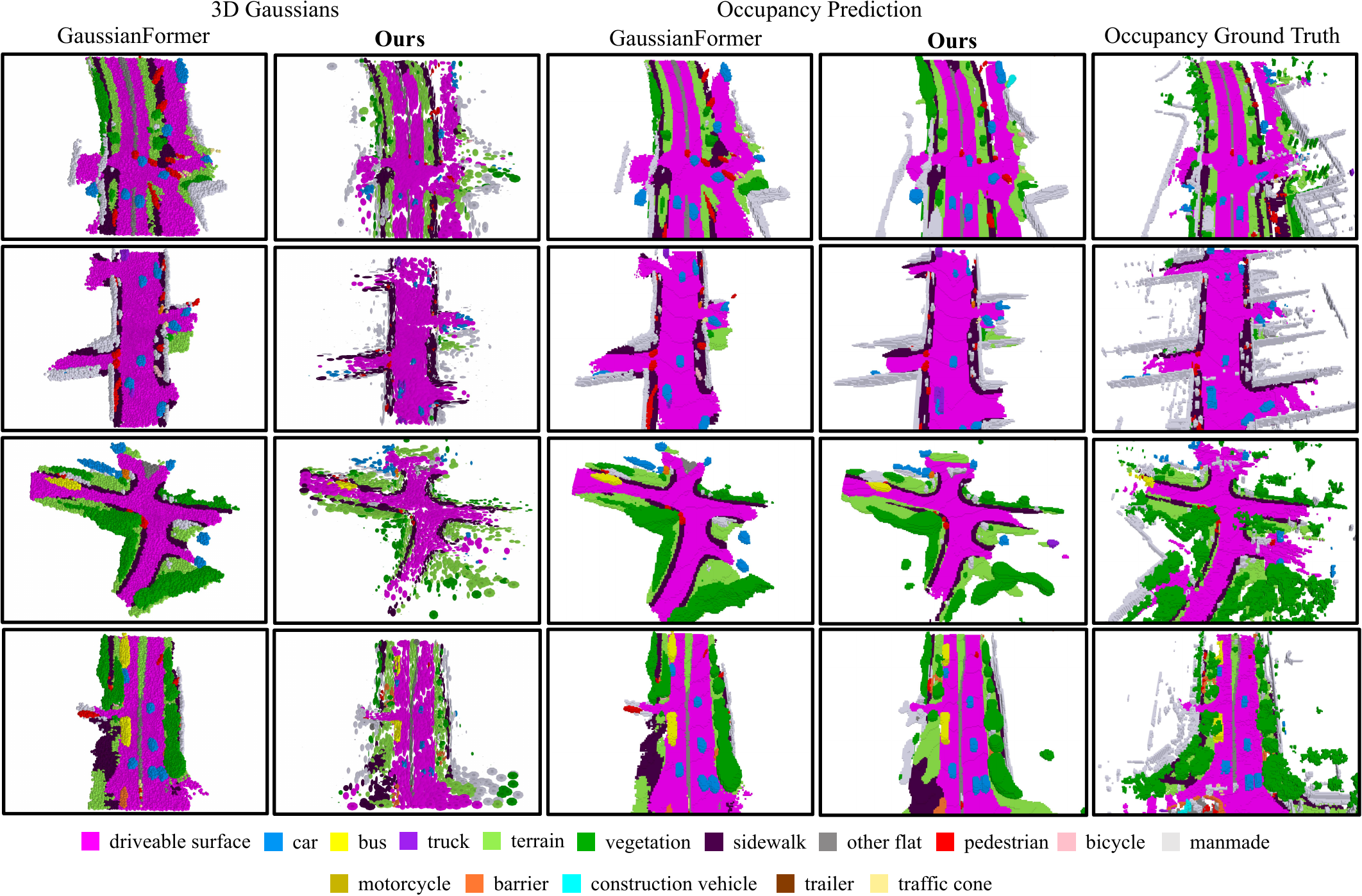}
\vspace{-2mm}
\caption{\textbf{Comparison with GaussianFormer~\cite{huang2024gaussian}.}
Our method predicts 3D Gaussians with more adaptive shapes compared with GaussianFormer.
Although our method uses only 8.8\% Gaussians, it still generates comprehensive occupancy predictions and alleviates the elongated effect in GaussianFormer.
}
\label{fig:comparison}
\vspace{-3mm}
\end{figure*}

\begin{figure*}[!ht]
\centering
\includegraphics[width=0.95\textwidth]{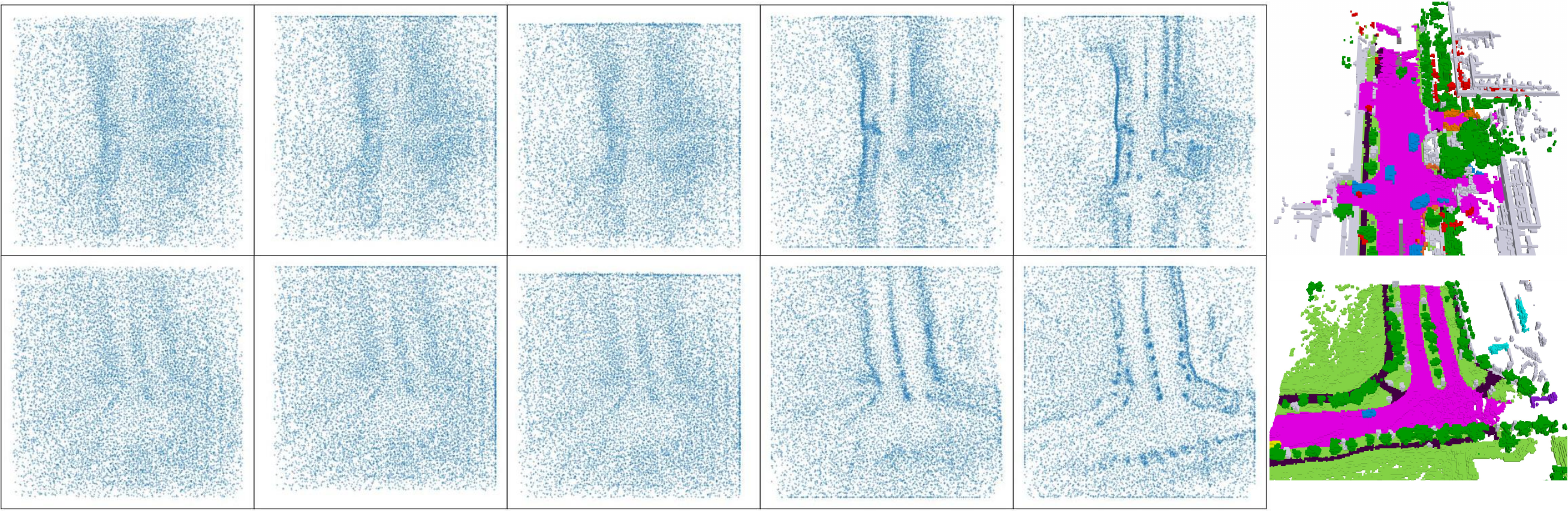}
\vspace{-2mm}
\caption{\textbf{Visualizations of Gaussian positions in the refinement process.}
We observe that our probabilistic Gaussians equipped with distribution-based intialization successfully learn to move towards occupied regions.
}
\label{fig:moving}
\vspace{-6mm}
\end{figure*}

\subsection{Ablation Study}

\textbf{Number of Gaussians.}
We report the influence of the number of Gaussians on the efficiency and performance of our model in Table~\ref{tab:number of gaussians}.
Our model achieves better performance-efficiency trade-off compared with GaussianFormer, outperforming it with less than 5\% number of Gaussians.
The latency of our method is higher than GaussianFormer, which we attribute to the time-consuming farthest point sampling (FPS) operation in our initialization module. 
We adopt the divide-and-conquer strategy to conduct the FPS operation in a batched manner for acceleration, and report the latency of the initialization module in parentheses.

\textbf{Design Choices.}
We conduct ablation study on the design choices of GaussianFormer-2 in Table~\ref{tab:components}.
We observe consistent improvement for both probabilistic modeling and distribution-based initialization module which surpasses the depth-based counterpart with a clear margin.

\textbf{Utilization of Gaussians.}
We provide comparisons on the utilization of Gaussians between GaussianFormer~\cite{huang2024gaussian} and our method in Table~\ref{tab:efficiency} using two important factors that reflect the utilization of Gaussians, which are position and overlap. 
Percentage of Gaussians in correct positions (Perc.) is percentage of Gaussians with their mean positions in the occupied space.
Overall overlap is calculated as summation of volumes of all Gaussians at 90\% over the coverage volume of all Gaussians, while individual overlap is computed by the average of the summation of the Bhattacharyya coefficient of each Gaussian with any other Gaussians.
We provide detailed information about these factors in the appendix.
Our method outperforms GaussianFormer on all these metrics, demonstrating better utilization.

\begin{table}[t]
    \centering
    \caption{
    \textbf{Ablation on the efficiency of GaussianFormer-2.}
    We set the number of Gaussians to 25600.
    Perc. and Dist. denote the percentage of Gaussians in correct positions, and the average distance of each Gaussian to its nearest occupied voxel, respectively.
    Overall and Indiv. represent the overall and individual overlapping ratios of Gaussians, respectively.
    }
    \vspace{-2mm}
    \setlength{\tabcolsep}{0.01\linewidth}
    \resizebox{1\linewidth}{!}{
    \begin{tabular}{c|cc|cc|cc}
        \toprule
        \multirow{2}{*}{Method} & \multicolumn{2}{c|}{Position} & \multicolumn{2}{c|}{Overlap} & \multirow{2}{*}{mIoU}   & \multirow{2}{*}{IoU} \\
        & Perc. (\%) \( \uparrow \) & Dist. (m) \( \downarrow \) & Overall \( \downarrow \) & Indiv. \( \downarrow \) \\
        \midrule
        GaussianFormer~\cite{huang2024gaussian} & 16.41 & 3.07 & 10.99 & 68.43 & 16.00 & 28.72   \\
        {Ours} & \textbf{28.85} & \textbf{1.24} & \textbf{3.91} & \textbf{12.48} & \textbf{20.32} & \textbf{31.04} \\
        \bottomrule
    \end{tabular}}
    \vspace{-7mm}
    \label{tab:efficiency}
\end{table}

\subsection{Visualizations}
We provide Gaussian and occupancy visualizations in Figure~\ref{fig:main}. 
Our model is able to predict reasonable Gaussian distributions and comprehensive occupancy results.
Further, we compare our method against GaussianFormer in Figure~\ref{fig:comparison}.
Our Gaussians are more adaptive in shape compared with isotropic spherical Gaussians in GaussianFormer.
We also visualize the xy coordinates of Gaussians in the initialization and subsequent blocks of GaussianFormer-2 in Figure~\ref{fig:moving}.
We find that the Gaussians successfully learn to move towards occupied area thanks to the object-centric probabilistic design and effective initialization module.

\section{Conclusion}
\label{sec: conclusion}

In this paper, we have proposed a probabilistic Gaussian superposition model as an efficient object-centric representation.
Specifically, we interpret each Gaussian as a probability distribution of its neighborhood being occupied and adopt the multiplication theorem of probability to derive the geometry predictions.
And we employ the Gaussian mixture model formulation to calculate normalized semantics predictions.
We have also designed a distribution-based initialization strategy to effectively initialize Gaussians around occupied area for object-centric modeling according to pixel-aligned occupancy distribution.
Our GaussianFormer-2 has achieved state-of-the-art performance on nuScenes and KITTI-360 datasets for 3D semantic occupancy prediction, which has also demonstrated improved efficiency compared with GaussianFormer on the number of Gaussians, position correctness and overlapping ratio.

\appendix

\begin{figure*}
    \centering
    \includegraphics[width=\linewidth]{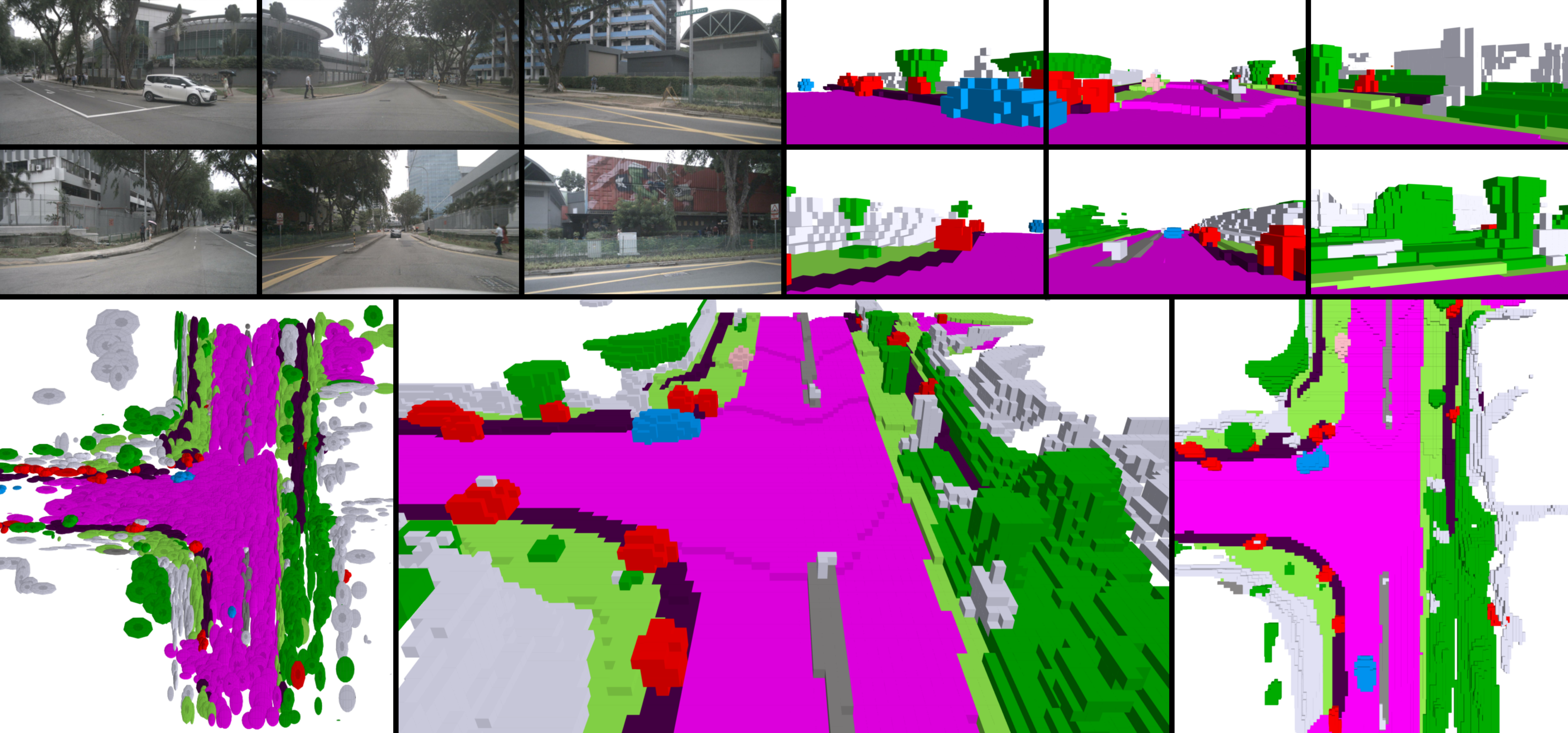}
    \vspace{-7mm}
    \captionof{figure}{
    \textbf{Visualizations of Gaussians, camera-view and overall occupancy on nuScenes.}
    We provide the input RGB images and their corresponding camera-view occupancy in the upper part.
    And we visualize the predicted 3D Gausians (left), the semantic occupancy in the global view (middle), and in the bird's eye view (right) in the lower part.
    }
\label{fig: supp teaser}
\end{figure*}%

\section{Video Demonstration}
\label{sec:video}
Figure~\ref{fig: supp teaser} shows a sampled frame of our video demonstration\footnote{\url{https://github.com/huang-yh/GaussianFormer}} for 3D semantic occupancy prediction on the nuScenes dataset~\cite{caesar2020nuscenes}.
We note that the camera-view occupancy visualizations align well with the input RGB images.
Moreover, each instance is sparsely described by only a few Gaussians with adaptive shapes, which demonstrates the efficiency and the object-centric nature of our model.

\section{Visualizations on KITTI-360}
\label{sec:vis kitti}
\begin{figure*}[t]
\centering
\includegraphics[width=0.95\linewidth]{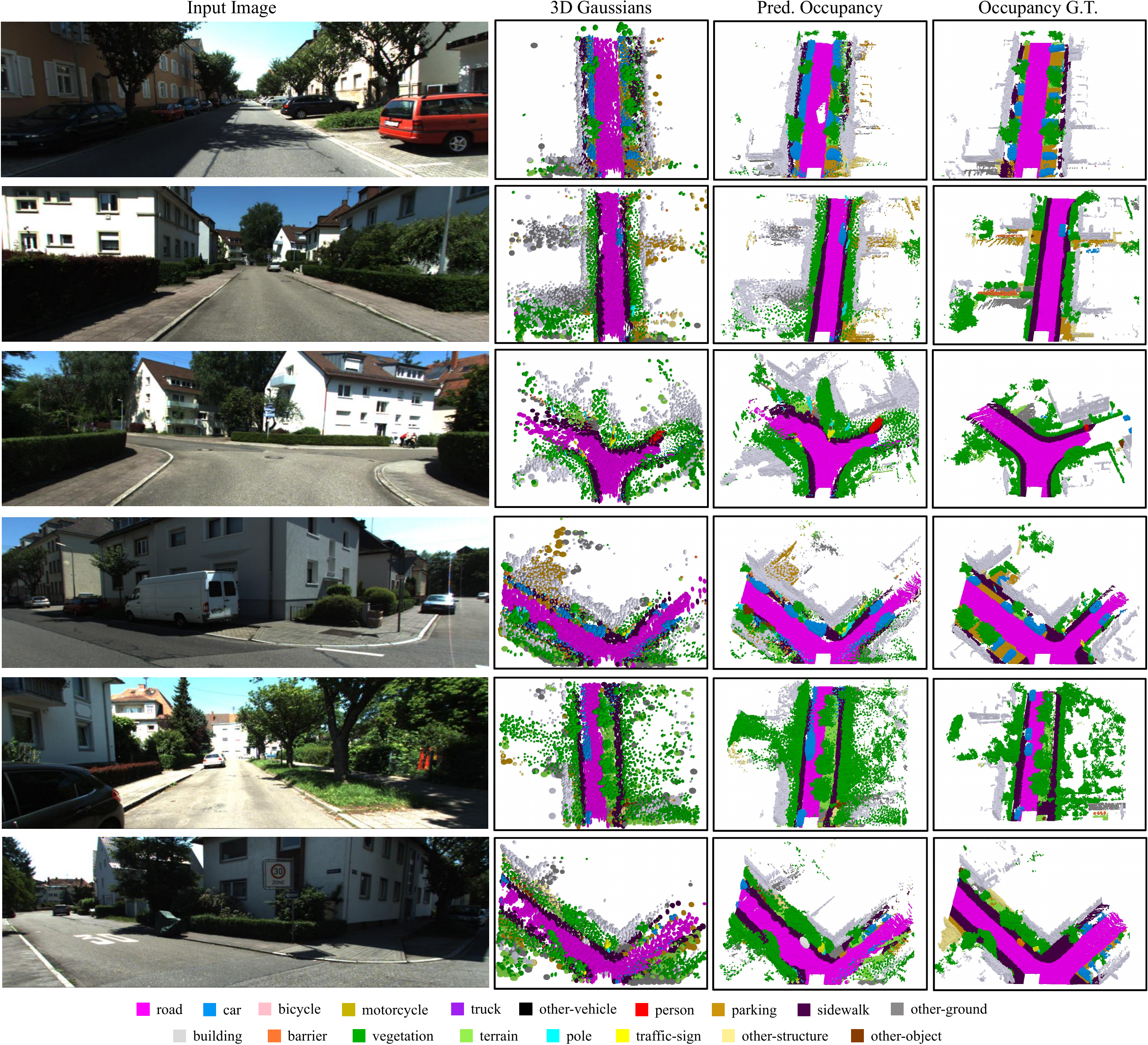}
\vspace{-2mm}
\caption{\textbf{Visualizations of Gaussians and occupancy on KITTI-360.}
Our method captures both the intricate geometry and semantics of the scene with shape-adaptive Gaussians.
}
\label{fig:supp kitti}
\vspace{-6mm}
\end{figure*}

We provide visualization results of Gaussians and occupancy on the KITTI-360 dataset~\cite{Liao2022kitti360} in Figure~\ref{fig:supp kitti}.
We observe that our GaussianFormer-2 is able to predict both intricate geometry and semantics of the 3D scene.
Furthermore, the 3D Gaussians in our model are adaptive in their scales according to the specific objects they are describing, compared with isotropic spherical Gaussians with maximum scales in GaussianFormer~\cite{huang2024gaussian}.

\section{Metric Details}
\label{sec:efficiency metrics}

\textbf{Position.} 
Gaussians, even after full training, can still be found in unoccupied space due to the localized nature of the receptive field. 
These Gaussians fail to describe meaningful structures, rendering them ineffective and devoid of practical utility. 
A higher proportion of Gaussians in unoccupied space indicates suboptimal utilization. 
Hence, we define the \textit{percentage of Gaussians in correct positions (Perc.)} as:
\begin{equation}
    {\rm Perc.} = \frac{N_{\text{correct}}}{N_{\text{total}}} \cdot 100\%,
    \label{eq: Perc.}
\end{equation}
where $N_{\text{correct}}$, and $N_{\text{total}}$ denote the number of Gaussians of which means are in occupied space, and the total number of Gaussians, respectively. 
A higher percentage indicates a better alignment of the Gaussians with occupied or meaningful area in the space, thus reflecting a more efficient use of the model's capacity.

The above measurement provides a hard evaluation, where Gaussians are either classified as being in correct or incorrect positions without considering their proximity to the nearest occupied area. 
This binary approach does not capture how close Gaussians in unoccupied regions are to meaningful positions. 
To address this limitation, we define a complementary soft measurement as the average distance of each Gaussian to its nearest occupied voxel center, denoted as \textit{Dist.} (in meters), computed as follows:
\begin{equation}
    {\rm Dist.} = \frac{1}{P} \sum_{i=1}^{P} \underset{\mathbf{v} \in \mathcal{V}}{\rm min} ||\mathbf{m}_i - \mathbf{v}||_1,
    \label{eq: Dist.}
\end{equation}
where $\mathbf{m}_i$, $\mathcal{V}$, $\mathbf{v}$, and $||\mathbf{m}_i - \mathbf{v}||_1$ denote the mean of the i-th Gaussian, the set of occupied voxel centers, the center of one voxel in this set, and L1 distance between the mean of the Gaussian and the voxel center, respectively.
Note that this distance is calculated with respect to the voxel center, and thus Gaussians positioned within the correct occupied area may also have a non-zero distance.

\textbf{Overlap.} 
The \textit{overall overlapping ratio of Gaussians (Overall.)} provides a global perspective on the redundancy in the Gaussian representation. 
Each Gaussian is modeled as an ellipsoid, where the semi-axis lengths are derived from the Mahalanobis distance at a chi-squared value of 6.251, corresponding to the 90\% confidence level of a Gaussian distribution in three degrees of freedom (DoFs).
The \textit{Overall.} is then calculated as the ratio of the summed 90\% confidence volumes $V_{i,90\%}$ of all Gaussians to the total coverage volume of all Gaussians $V_{\text{coverage}}$ in the scene:
\begin{equation}
    {\rm Overall.} = \frac{\sum_{i=1}^{P} V_{i,90\%}}{V_{\text{coverage}}}, 
    \label{eq: Overall.}
\end{equation}
where $V_{\text{coverage}}$ represents the volume of all Gaussians combined as a unified shape. To estimate $V_{\text{coverage}}$, we employ the \textit{Monte Carlo method} where a large number of points are randomly sampled within the bounding box of the scene. For each sampled point, we check whether it lies within the 90\% confidence ellipsoid of any Gaussian. The volume is then approximated as:
\begin{equation}
    V_{\text{coverage}} = V_{\text{scene}} \cdot \frac{N_{\text{in}}}{N_{\text{total}}},
    \label{eq: Monte}
\end{equation}where $N_{\text{in}}$, and $N_{\text{total}}$ are the number of sampled points that fall within the 90\% confidence ellipsoid of at least one Gaussian, and the total number of sampled points, respectively. This approach ensures an accurate estimation of the unified volume, efficiently handling the overlapping regions of the Gaussians by not double-counting them.

We next detail the derivation of the ellipsoid volume corresponding to the 90\% confidence region of a 3D Gaussian distribution. 
Considering a multivariate Gaussian distribution in 3D defined as:

\vspace{-4mm}
\begin{small}
\begin{equation}
    \mathbf{g}(\mathbf{x}) = \frac{1}{(2 \pi)^{3/2}|\mathbf{\Sigma}|^{1/2}}{\rm{exp}}\big(-\frac{1}{2}(\mathbf{x}-\mathbf{m})^{\rm T} \mathbf{\Sigma}^{-1} (\mathbf{x}-\mathbf{m})\big),
    \label{eq: gaussian_dist}
\end{equation}
\end{small}where $\mathbf{x}$, $\mathbf{\Sigma}$, and $|\mathbf{\Sigma}|$ are the mean vector, 3x3 covariance matrix, and the determinant of the covariance matrix, respectively.
The \textit{Mahalanobis distance} $d$ of point $\mathbf{x}$ from the mean $\mathbf{m}$ is defined as:
\begin{equation}
    d^2(\mathbf{x},\mathbf{m}) = (\mathbf{x}-\mathbf{m})^{\rm T} \mathbf{\Sigma}^{-1} (\mathbf{x}-\mathbf{m}).
    \label{eq: maha_dist}
\end{equation}
The 90\% confidence region of the Gaussian distribution corresponds to the set of points for which the Mahalanobis distance satisfies:
\begin{equation}
    d^2 \le \chi_{3,0.9}^2 \approx 6.251,
    \label{eq: chi2}
\end{equation}
where $\chi_{3,0.9}^2$ is the chi-square critical value for three degrees of freedom at the 90\% confidence level.
For a Gaussian distribution, the semi-axis lengths are determined by the square root of the eigenvalues of $\mathbf{\Sigma}$, scaled by $\chi_{3,0.9}^2$.
Thus, the volume of the ellipsoid from 90\% of the 3D Gaussian distribution is:
\begin{equation}
    V_{90\%} = \frac{4}{3}\pi (6.251)^{3/2} |\mathbf{\Sigma}|^{1/2}.
    \label{eq: V_2}
\end{equation}
A higher value of \textit{Overall.} indicates greater overlapping volumes among the Gaussians, signifying redundancy in Gaussian representation. 
This metric provides insights into the utilization of Gaussians to represent the scene.

The \textit{individual overlapping ratio of Gaussians (Indiv.)} offers a fine-grained analysis of the overlap between Gaussians in a scene. 
This measurement quantifies the degree to which each Gaussian overlaps with all other Gaussians, averaged across all Gaussians in the scene. 
The value of this metric indicates approximately how many times the volume of a single Gaussian is fully overlapped with other Gaussians on average.
To compute this, we use the Bhattacharyya coefficient~\cite{bhattacharyya1943bhatcoef}, which measures the similarity between two Gaussian distributions. 
The \textit{individual overlapping ratio} is defined as:
\begin{equation}
   {\rm Indiv.} = \frac{1}{P}\sum_{i=1}^{P}\left(\sum_{j\ne i}{\rm BC}_{i,j}\right),
    \label{eq: indiv}
\end{equation}
where ${\rm BC}_{i,j}$ is the Bhattacharyya coefficient between the i-th and j-th Gaussians, given by:
\begin{equation}
   {\rm BC}_{i,j} = \frac{\sqrt[4]{|\mathbf{\Sigma}_i||\mathbf{\Sigma}_j|}}{\sqrt{|\mathbf{\Sigma}_{ij}|}}e^{-\frac{1}{8}(\mathbf{m}_i-\mathbf{m}_j)^{\rm T}\mathbf{\Sigma}_{ij}^{-1}(\mathbf{m}_i-\mathbf{m}_j)},
   \label{eq: Batta}
\end{equation}
where $\mathbf{\Sigma}_{ij}=\frac{1}{2}(\mathbf{\Sigma}_i+\mathbf{\Sigma}_j)$ is the average covariance matrix.
A higher value of \textit{Indiv.} indicates more redundancy, as Gaussians are heavily overlapping with each other.

\end{document}